# Attention-Sensitive Alerting

Eric Horvitz, Andy Jacobs, David Hovel
Microsoft Research
Redmond, Washington 98025
{horvitz,andyj,davidhov}@microsoft.com

## Abstract

We introduce utility-directed procedures for mediating the flow of potentially distracting alerts and communications to computer users. We present models and inference procedures that balance the context-sensitive costs of deferring alerts with the cost of interruption. We describe the challenge of reasoning about such costs under uncertainty via an analysis of user activity and the content of notifications. After introducing principles of attention-sensitive alerting, we focus on the problem of guiding alerts about email messages. We dwell on the problem of inferring the expected criticality of email and discuss work on the PRIORITIES system, centering on prioritizing email by criticality and modulating the communication of notifications to users about the presence and nature of incoming email.

## 1 Introduction

Multitasking computer systems provide great value to users by hosting numerous processes and applications simultaneously. However, the ongoing execution of multiple applications often leads to environments fraught with a variety of notifications, including messages from the operating system about the status and health of computational processes, alerts from the primary application at focus, and from other applications being executed in the background.

Beyond traditional sources of peripheral information, recent work on human–computer interaction highlights new forms of ongoing background services that can provide potentially useful context-sensitive information and analysis (Breese, Heckerman, & Kadie, 1998; Czerwinski, Dumais, & Robertson et al.; Leiberman, 1995; Horvitz, Breese, Heckerman et al., 1998; Horvitz, 1999). Indeed, novel sources of information, as well as more familiar alerts about incoming email messages, tips about application usage, and information about the computer system and network may be valuable. However, the rendering of auxiliary information under uncertainty comes at the cost of potentially distracting the user from a primary task at the focus of attention.

We are exploring utility-directed notification policies within the *Attentional Systems* project at Microsoft Research. We shall describe procedures that can provide policies to support an automated *attention manager* that one day might be relied upon by computer users to mediate the transmission of notifications.

We take the perspective that human attention is the most valuable and scarcest commodity in human–computer interaction. Rapid increases over the last two decades in computational power and network bandwidth, coupled with the explosion in the availability of online content, stand in stark contrast to the constancy of limitations in human information processing.

Characterizations of the inability of people to handle more than a handful of concepts in the short-term are perhaps the most critical results of Twentieth-century psychology (Miller, 1956; Waugh, 1965). Beyond general characterizations of cognitive limitations, psychologists have explored the influence of various forms of interruption on human memory and planning, starting with the early work of Zeigarnik and Ovsiankina (Zeigarnik, 1927; Ovsiankina, 1928). The rich body of work in this realm includes studies centering on the use of interruptions as a tool to probe the machinery of memory and problem solving as well as to ascertain the influence of distractions on the efficiency with which tasks are accomplished (Gillie & Broadbent, 1989; Van Bergan, 1968; Posner & Konick, 1966).

We have been pursuing opportunities to harness inference and decision-making procedures to guide the rendering of notifications about messages of uncer-tain value. Our approach centers on developing the means for automatically assessing the expected utility of messages and for continuing to make inferences about a user's focus of attention by monitoring multiple sources of information.

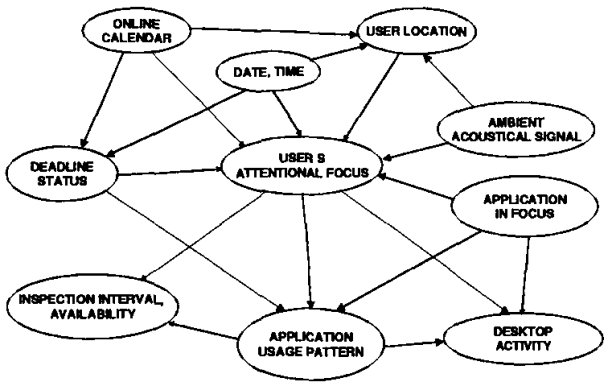

Figure 1: A Bayesian model for inferring the probability distribution over a user's attentional focus.

We shall focus first on the use of Bayesian models to infer a probability distribution over a user's focus of attention and harnessing such inferences to infer the expected cost of transmitting alerts to users. Then, we consider methods for inferring the informational benefits of alerts and the costs of deferring notification. After discussing principles of alerting based on a consideration of probability distributions over a user's attention and the time criticality of alerts, we shall present selected details of work on developing notification and forwarding policies for incoming email.

## 2  Inference about a User's Attention

Alerts provide potentially valuable information at a cost of interruption. The cost of an interruption depends on the nature of the interruption and on a user's current task and focus of attention. In the general case, a computer system is uncertain about the details of a user's attention. Thus, we seek to build or learn probabilistic models that can make inferences about a user's attention under uncertainty.

We have pursued the construction of Bayesian models that can infer a probability distribution over a user's focus of attention. In building probabilistic models for inferring the context-sensitive cost of distraction, we consider a set of mutually exclusive and exhaustive states of attentional focus and seek to identify the cost of communicating an alert given a probability distribution over the states of a user's attention. Such states of attention can be formulated as a set of prototypical situations or more abstract representations of a set of distinct classes of cognitive challenges being addressed by a user. Alternatively, we can formulate models that make inferences about a continuous measure of attentional focus, or models that directly infer a probability distribution over the cost of interruption for different types of notifications. In our initial approach to modeling a user's attention, we have Bayesian networks that can be used to infer the probability of alternate activity contexts based on a set of observations about a user's activity and location.

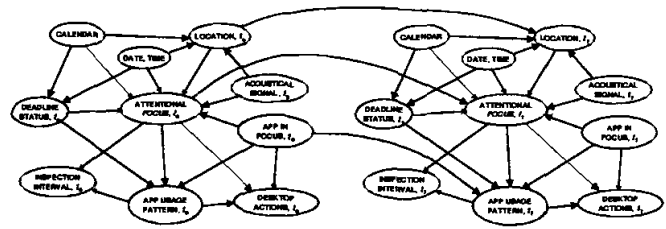

Figure 2: Extending the Bayesian network to consider key dependencies over time.

Figure 1 displays a Bayesian network for inferring a user's focus of attention for a single time period. States of the critical variable, FOCUS OF ATTENTION, refer to desktop and nondesktop contexts. Sample attentional contexts considered in the model include SITUATION AWARENESS–CATCHING UP, NONSPECIFIC BACKGROUND TASKS, FOCUSED CONTENT GENERATION OR REVIEW, LIGHT CONTENT GENERATION OR REVIEW, BROWSING DOCUMENTS, MEETING IN OFFICE, MEETING OUT OF OFFICE, LISTENING TO PRESENTATION, PRIVATE TIME, FAMILY–PERSONAL FOCUS, CASUAL CONVERSATION, and TRAVEL.

The Bayesian network specifies that a user's current attention and location are influenced by the user's scheduled appointments, the time of day, and the proximity of deadlines. The probability distribution over a user's attention is also influenced by summaries of the status of ambient acoustical signals monitored in a user's office; segments of the ambient acoustical signal over time provide clues about the presence of activity and conversation. The status and configuration of software applications and the ongoing stream of user activity generated by a user interacting with a computer also provide rich sources of evidence about a user's attention. As portrayed in the network, the software application currently at top-level focus in the operating system influences the nature of the user's focus and task, and the status of a user's attention and the application at focus together influence the computer-centric activities. Such activity includes the stream of user activity built from sequences of mouse and keyboard actions (see Horvitz, Breese, Heckerman et al., 1998 for a discussion of events and event languages for monitoring user behavior) and higher-level patterns of application usage over broader time horizons. Such patterns include EMAIL-CENTRIC and WORD-PROCESSOR CENTRIC, referring to prototypical classes of activity involving the way multiple applications are interleaved.

A more comprehensive Bayesian model for a user's attentional focus considers key dependencies among variables at different periods of time. A dynamic network model including a set of Markov temporal dependencies is portrayed in Figure 2. In real-time use, such Bayesian models consider information provided by an online calendar, and a stream of observations about room acoustics and user activity as reported by an event sensing system, and continues to provide inferential results about the probability distribution a user's attention.

## 3 Expected Cost of Interruption

Let us assume that the expected utility of relaying information contained in an alert to a user can be decomposed into the expected costs and benefits of the alerting action. For such decomposable utility models, we can assume that the utility is the difference between the expected costs and benefits of the information provided by the alert. We focus first on the expected cost of immediate alerting.

Alerts and notifications can take the form of audio, visual, or a combination of audio and visual channels. Beyond the cognitive cost of the immediate distraction associated with an alert, visual alerts can obstruct important content being accessed or referred to as part of the task at hand. The cost associated with an autonomous notification can depend on the details of the rendering of the alert. Thus, in the general case, distinct dimensions of cost associated with different notification designs must be considered in models of interruption.

As an example, it may be useful to decompose the cost of an alert into the cognitive cost associated with an interruption and the cost of obstruction of important display real estate. The latter dimension of cost can depend significantly on the design of the visual alert and the status of displayed information associated with the main task at hand.

A design that overlays a graphical notification over content at the center of a user's attention and that requires a user to take action to remove the displayed alert is more costly than an alert that appears and disappears autonomously in a timely and elegant manner. For simplification, we shall merge the cost of interruption and the cost of obstruction into a single cost. The generality of the methods will not suffer from such a coalescence.

Consider a set of alerting outcomes, $A_i, F_j$, representing the situation where a notification $A_i$ occurs when a user is in a state of attentional focus, $F_j$. We assess for each alerting outcome, a cost function of the form $C^a(A_i, F_j)$, referring to the cost of being alerted via action $A_i$ when the user is in attentional state $F_j$. Given uncertainty about a user's state of attention, the expected cost of alerting (ECA) a user with action $A_i$ is,

$$\text{ECA} = \sum_j C^a(A_i, F_j) p(F_j | E^a) \qquad (1)$$

where $E^a$ refers to evidence relevant to inferring a user's attention.

## 4 Expected Cost of Deferring Alerts

A strategy for reducing the cost associated with alerts is to suppress the alerts or to defer them until a period of time when the cost of relaying them is smaller. Decisions about deferral must take into consideration the cost associated with the delayed review of the information. We now turn to the expected cost associated with deferring the review of a notification for some time $t$.

### 4.1 Cost of Delayed Action

We define the *criticality* of a notification as the expected cost of delayed action associated with reviewing the message. The expected cost of delayed action (ECDA) has been applied in such domains as emergency medicine (Horvitz & Rutledge, 1991; Horvitz & Seiver, 1997) and time-critical aerospace applications (Horvitz & Barry, 1995). ECDA is the difference in the expected value of taking immediate ideal action (action at time $t_o$), and delaying the ideal action until some future time $t$. Given a probability distribution, $p(H|E)$, over states of the world $H$, associated with different time criticalities, and a time-dependent utility function over outcomes, $u(A_i, H_j, t)$, the expected cost of delayed action for notifications is,

$$\text{ECDA} =$$
$$\max_A \sum_j u(A_i, H_j, t_o) p(H_j | E)$$
$$- \max_A \sum_j u(A_i, H_j, t) p(H_j | E) \qquad (2)$$

ECDA provides a conceptual framework for reasoning about the cost of the delayed review of notifications.

Let us consider the example of decisions about notifying users about the arrival of messages via email.

We must consider the criticality of the email and the cost of interruption associated with the user's focus of attention. Notification about email includes desktop alerting when the user is working at or near a computer and notification via a mobile communication device, such as a cell phone or pager, when the user is away from a networked computer.

The utility of reading an email message can diminish significantly with delay in reviewing the message. In a salient example, delay in reviewing a message that informs a user about a competitive bidding situation can lead to a costly loss of opportunity. Costs of delayed review of messages may be high in the context of communications involving coordination. Important meetings and deadlines can be missed with delayed review of messages. In less severe situations, costs can accrue with reductions in the amount of time available to prepare effectively for a meeting. For such cases, the cost of delayed review of messages can be represented by loss functions that operate on the amount of time remaining until the meeting being communicated about occurs. After a meeting has passed, many options for action are eliminated. Thus, the rate of loss incurred with delays in the review of a message are typically smaller for periods of time following the occurrence of a meeting described in an email message.

We could attempt to group messages into classes indexed by the types of action indicated at progressively later times and endeavor to formulate a set of outcomes associated with ideal actions at different delays in reviewing the messages. With such a representation, Equation 2 could be used to compute an expected cost of delayed review directly. Alternatively, we can simplify ECDA by considering the probability that a message is a member of one of several criticality classes, given features of the messages. We associate with each criticality class a time-dependent cost function, describing the rate at which losses accrue with delayed review of the message. We take $t_o$ to be the moment that email arrives and compute the expected cost for delays in reviewing the message until time $t$. In the general case, the costs of delayed review for messages in each criticality class may be a nonlinear function of delayed review.

The complexity and scope of communications among people makes the certain identification of the criticality of email messages difficult. It is more feasible to pursue inference about a probability distribution over the criticality of a message given evidence gleaned from attributes of the message, including information contained in the header and body of email messages.

We shall return to explore in detail methods for learning the criticality of email messages in Section 5. For now, let us assume that each message is a member of one of $n$ criticality classes. We further assume that each class is associated with a criticality-class–specific constant rate of loss that describes the cost of delayed review. Using $C^d$ to represent a time-dependent rate of loss with delay, we can reduce Equation 2 to an expected cost of delayed review (ECDR),

$$\text{ECDR} = \sum_i (t - t_o) C^d(H_i) p(H_i | E^d) \qquad (3)$$

where $t_o$ represents the time a message arrives, $t$ is the time the message is reviewed, and $E^d$ is evidence used to infer a probability distribution over the criticality class, $H$, of a new incoming message at hand. We refer to the constant rate of loss associated with delayed review as the expected criticality (EC) of a message,

$$\text{EC} = \sum_i C^d(H_i) p(H_i | E^d) \qquad (4)$$

### 4.2 Ideal Alerting about New Messages

Users typically review email periodically even when their computing systems are configured to suppress the active emission of alerts about incoming email. To develop ideal alerting policies, we consider the cost of delayed review of information incurred in a world absent of notifications. The cost of delay in such settings depends on the criticality of the message and the time passing before a user reviews a message without external prompting.

The expected delay in the review of messages in an alert-free setting can be inferred from information about the frequency that users will attend to unread messages without prompting. Beyond considering the frequency that users will review messages on their own, we can consider expected delays associated with a policy of relaying notifications to users when the cost of interruption is inferred to be negligible. With such a policy in force, the delay until new messages are reviewed can be inferred from information about the expected time before a user's attentional resources will become freed to review the messages.

We refer to the time between periods of reviewing new messages in the absence of explicit alerts as the *inspection interval*, $I$. The inspection interval is influenced by multiple factors including the user's focus of attention and location. A user's inspection interval is typically reduced when they are at a distance from networked computers.

The Bayesian networks presented in Figure 1 and 2 include a variable representing the inspection interval. As displayed by the dependency structure of the models, the variable INSPECTION INTERVAL, is influenced by USER'S ATTENTIONAL FOCUS and APPLICATION USAGE PATTERN.

Given a probability distribution over the inspection interval, the expected loss associated with reviewing messages in an alert-free setting, ECDR', is

$$\text{ECDR}' = \sum_j p(I_j)(t_{I_{-1}} + I_j - t_o) \sum_i C^d(H_i) p(H_i|E^d) \quad (5)$$

where $t_{I_{-1}}$ is the time of last access, $t_o$ is the time a message has arrived, and $I_j$ is the inspection interval.

The expected value of transmitting an alert (EVTA) about a message at some time $t$ before a user reviews the email is the increase in the expected utility with being informed about the message at $t$ versus at the time we expect the user to access the email in the absence of an alert. That is,

$$\text{EVTA} = \sum_j p(I_j)(t_{I_{-1}} + I_j - t_o) \sum_i C^d(H_i) p(H_i|E^d) \\ - \sum_i (t - t_o) C^d(H_i) p(H_i|E^d) \quad (6)$$

A system should relay information about a message when the net value of the alert (NEVA) is positive. This is the case when the EVTA dominates the immediate ECA for the type of alert under consideration,

$$\text{NEVA} = \text{EVTA} - \text{ECA} \quad (7)$$

### 4.3 Chunking Messages and the Value of Alerting

The grouping together of information from multiple messages into a single compound alert can raise the value of the content revealed under the guise of a single, but potentially more complex, distraction. Reviewing information about multiple messages in an alert can be more costly than an alert relaying information about a single message. We represent such increases in distraction by allowing the cost of an alert to be a function of its informational complexity.

Let us assume that the EVA of an email message is independent of the EVA of other email messages. We use $\text{EVTA}(M_i, t)$ to refer to the value of alerting a user about a single message $M_i$ at time $t$ and $\text{ECA}(n)$ the expected cost of associated with relaying the content of $n$ messages. We can modify Equation 7 to consider multiple messages by summing together the expected value of relaying information about a set of $n$ new messages,

$$\text{NEVA} = \sum_{i=1}^n \text{EVTA}(M_i, t) - \text{ECA}(n) \quad (8)$$

We note that assuming independence in the value of reading distinct messages may lead to an overestimation of the value of the multiple-message alert because strings of messages received in sequence may refer to related content.

Given inferred probability distributions over a user's attentional focus and inspection interval, an assessment of the costs of distracting a user with alerts, and the time criticality of incoming messages, we can employ NEVA to continue to reason about the costs versus the benefits of alerting users with summarizing information about the content of newly arriving email messages. We now turn to the task of automatically assigning measures of expected criticality to email messages.

## 5 Assigning Criticality to Messages

Building a real-world system for exploiting NEVA to control alerting hinges on an ability to automatically assign a measure of expected criticality to incoming messages. Given the challenge and importance of making inferences about the criticality of alerts, we shall dwell on details of inferring the expected criticality of email messages. Such methods have application to other classes of notifications.

We have developed an automated criticality classifier for email by leveraging and extending learning and inference methods developed for performing text classification. The methodology employs several phases of analysis including: (1) selection of features, (2) construction of a classifier, (3) mapping classifier outputs to the likelihood that an email message is a member of each criticality class, and (4) the computation of an expected criticality from the probability distribution over criticality classes for email messages.

Text classification is an active area of research and development (see Dumais, Platt, Heckerman et al, 1998 for a review of recent efforts. Machine learning methods employed in text classification include decision trees (Lewis & Ringuette, 1994), regression (Yang & Chute, 1994), Bayesian models (Lewis & Ringuette, 1994; Sahami, 1996; Sahami, Dumais, Heckerman et al., 1998), and Support Vector Machines (Joachims, 1998; Scholkopf, Burges, & Smola, 1998).

Our group has been studying the characteristics and performance of several text classification methodologies for classifying email including procedures based on Bayesian network learning procedures (Sahami, Dumais, Heckerman et al., 1998) and the Support Vector Machine learning methodology (Vapnik, 1995; Platt, 1999a). Our studies of standard test corpora (e.g., Reuters corpora of business articles) and a variety of text classification tasks demonstrated that specific forms of SVM strategies dominated naive Bayes classification procedures developed to date for text classification (Dumais, Platt, Heckerman et al, 1998).

Our current implementation of criticality assignment for email is based on a linear Support Vector Machine training methodology developed by Platt called Sequential Minimal Optimization (Platt, 1999a). Support Vector Machines build classifiers by identifying a hyperplane that separates a set of positive and negative examples with a maximum margin (see Platt, 1999a for details). In the linear form of SVM that we employ to assign criticality classes to email, the margin is defined by the distance of the hyperplane to the nearest positive and negative cases for each class. Maximizing the margin can be expressed as an optimization problem and search and optimization thus lay at the core of different SVM-based training methods.

Traditionally, SVM training methods yield classifiers that output a score describing the strength of membership in a category. Platt has extended SVM methods by developing a methodology that provides an estimate of the probabilities that items are members of different classes (Platt, 1999b). The procedure employs regularized maximum likelihood fitting to produce estimations of posterior probabilities. We harnessed this approach to learn classifiers that output the probability that an email message is a member of different criticality classes.

In practice, we create a set of criticality classes and assess time-dependent cost functions for each class. We obtain a training set by manually partitioning a corpus of sample messages into distinct criticality classes. Given a training corpus of messages labeled by criticality, we first apply feature-selection procedures that attempt to find the most discriminatory features for the set of target classes, using several phases of analysis including a mutual-information analysis (Koller & Sahami, 1996; Dumais, Platt, Heckerman et al., 1998; Sahami, Dumais, Heckerman et al., 1998). We refer readers to the text-classification literature for details on practical and theoretical issues in feature selection.

Feature selection procedures for text classification can operate on single words or higher-level distinctions made available to the algorithms, such as phrases and parts of speech tagged with natural language processing. Basic feature selection algorithms for text classification typically perform a search over single words. Beyond the reliance on single words, we can make available to feature selection procedures domain-specific phrases and high-level patterns of features, including general expressions that operate on classes of words and other features in email messages. We found that providing such special tokens to text-classification procedures can enhance classification significantly(Sahami, Dumais, Heckerman et al., 1998 ).

In investigating the construction of classifiers for email criticality, we identified special phrases and other classes of observations that we suspected could be of value for discriminating among email messages associated with different time criticalities. The handcrafted features are considered during feature selection. Tokens and patterns of value in identifying the criticality of messages include such distinctions as:

- Sender: Single person versus an email alias, people at a user's organization, organizational relationship to user, names included on a user constructed list, people user has replied to

- Recipients: Sent only to user, sent to a small number of people, sent to a mailing list

- Time criticality: Inferred time of an implied meeting, language indicating cost with delay, including such phrases as "happening soon," "right away," "as soon as possible," "need this soon," "right away," "deadline is" "by time, date," etc.

- Past tense: Phrases used to refer to events that have occurred in the past such as, "we met," "meeting went," "took care of," "meeting yesterday," etc.

- Future tense: Phrases used to refer to events that will occur in the future including "this week," "Are you going to," "when are you," etc.

- Future dates: Days and times representing future dates.

- Coordination: Language used to refer to coordinative tasks such as "get together," "can we meet," "coordinate with," etc.

- Personal requests: Phrases associated with direct requests for assistance, including sentences ending with question marks, "will you," "are you," "can you," "I need," "take care of," "need to know," etc.

- Importance: Language and symbols referring to importance including the presence of an explicit high or low priority flag, and such phrases as "is important," "is critical," etc.

- Length of message: Size of new component of a message (excluding the forwarded thread)

- Presence of attachments: Noting the inclusion of documents in the email

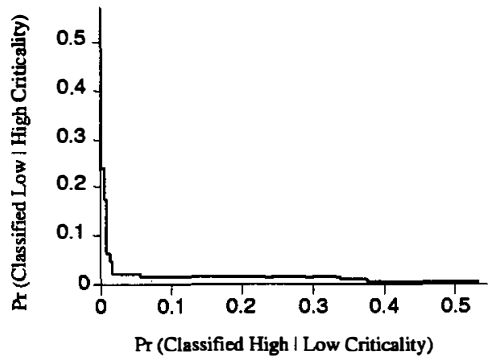

Figure 3: Discriminatory power of an email criticality classifier. The curve indicates the probability of misclassification at different decision thresholds for a test set of hand selected messages in high and low criticality classes.

- Time of day: The time a message was composed.

- Signs of Junk email: patterns such as percent nonalphanumeric characters, and pornographic content, marketing phraseology such as "Free!," "Only $," "Limited offer," etc.

We found that the coupling of an SVM classifier with criticality-specific tokens can effectively classify email into criticality classes and into overall estimates of expected criticality. In an evaluation, a criticality classifier was trained from approximately 1500 messages, divided into approximately equal sets of low and high priority email messages. A curve showing the ability of the classifier to classify messages from a test corpora consisting of 250 high and 250 low priority messages, selected by a user from a large inbox, is displayed in Figure 3. The Receiver-Operator (ROC) curve displays the probability of high priority email being classified as low priority email and the probability of low priority email being classified as high priority email for different values of the probability threshold used to define the high and low criticality message classes.

Although it is useful to demonstrate the ability of the classifier to appropriately label cases of low and high criticality email, we are most interested in the use of the inferred probabilities of membership in alternate classes to compute the expected criticality of messages, and in the ultimate use of such information in computing the NEVA associated with messages.

As part of the validation of the automated assignment of measures of criticality for email, we generated expected criticalities of email messages, assuming a linear cost of delay with time for each criticality class, and summing the costs for each class weighted by the probability that messages are members of each class as reported by the classifier. Our validations have shown that the classifier performs well even with the use of only two classes of criticality: time-critical messages and normal/low priority messages. In a validation study, one of the authors scored the criticality of messages by hand on a 1 to 100 scale, using 1 to indicate the messages of lowest criticality and 100 to represent the most time-critical messages. To probe the effectiveness of the expected criticality measure, we computed correlation coefficients and generated scatter plots to visualize relationships between the assessed criticalities and the computed expected criticality. In a sample study, one of the authors assessed the criticality of 200 email messages received over three days. A correlation coefficient of 0.9 was found between the user tagged criticality and the automated assignment of expected criticality.

## 6 PRIORITIES Prototypes

We have been exploring the use of attention management for email messages through implementations of several prototypes we refer to as the PRIORITIES family of systems. The PRIORITIES prototypes learn classifiers from examples drawn from a user's email and apply the classifiers in real time to assign expected criticalities to incoming email messages. The systems work with the MS Outlook 2000 messaging and calendar system. During feature selection, the systems consider categories of features described in Section 5.

The classification learning and inference procedures have been integrated in a software application that calls the Microsoft Exchange MAPI and Outlook 2000 CDO interfaces. These services grant the system access to details of the message header, including sender and recipient information, and the organizational hierarchy at Microsoft. When email arrives, the real-time classifier examines the incoming messages for words and phrases and makes calls to acquire sender, recipient, and organizational information.

An early version of PRIORITIES has been distributed widely at Microsoft for real-world testing. This version assigns a measure of expected criticality to all incoming mail, using a pretrained, default classifier or a classifier that is custom-trained by the onboard learning subsystem. The system has been integrated with the MS Research EVE event sensing system, developed as part of the LUMIÈRE intelligent interface project (Horvitz, Breese, & Heckerman et al., 1998), enabling the system to continue to consider a variety of observations, including keyboard and mouse activity, and room acoustics. Information about a user's schedule is accessed directly from Outlook's online calendar.

The version of the PRIORITIES system that is currently being tested by users at Microsoft provides an email viewer client that displays email sorted by criticality and scoped by a user-specified period of time. A display of the *Priorities* client is displayed in Figure 4. The prototype can be instructed to take a variety of actions based on observations about the user's activity and location, and the inferred expected criticality of incoming mail. Actions include playing criticality-specific sounds that were specially composed for the system, bringing the client to the foreground, and opening email messages and sizing and centering the email according to criticality. The system can be directed to perform a variety of automated forwarding and response services based on expected criticality. Moving beyond the desktop, the system has the ability to forward messages to a user's cell phone or pager based on criticality and the time a user is away from the office. For mobile settings associated with limited time and bandwidth, PRIORITIES can be employed to download messages in order of expected criticality.

A more advanced version of PRIORITIES, named PRIORITIES–ATTEND serves as our testbed for performing more sophisticated inference about a user's attention and for making decisions about notification based on NEVA. This version has been integrated with a manually constructed Bayesian network that performs inference about a user's attention. Work is underway on the development of effective assessment techniques and richer models for representing and reasoning about a user's attention and the costs of interruption. Our experiences to date with the use of automated alerting machinery suggest that a decision-theoretic approach to alerting can fundamentally change the way users work with email communications.

## 7 Summary

We have described efforts to harness decision-theoretic principles to control alerting in computing and communication systems. We presented attention-sensitive procedures for computing the net expected value of alerts. We framed the discussion with the task of relaying notifications about incoming email messages. After presenting principles for decisions about alerting users about messages, we presented work on automatically assessing the expected criticality of email messages. Finally, we presented work on the PRIORITIES systems, prototypes that operate with the Microsoft Outlook email and scheduling application.

There are numerous opportunities for enhancing the value of computing systems through harnessing methods that perform ongoing inference about a user's at-tention and about the criticality of different sources of information. We are continuing our pursuit of decision-theoretic machinery that can endow operating systems with the ability to monitor multiple sources of information and make intelligent decisions about the expected value of transmitting notifications to users.

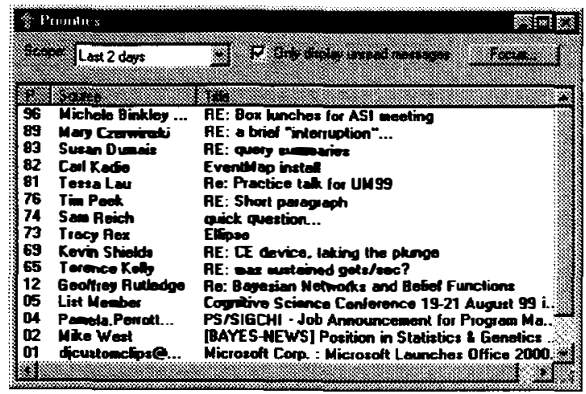

Figure 4: Display provided by the client of a version of PRIORITIES being tested by multiple users. The client comes into view upon demand or when criticality-directed policies bring it to the foreground.

## Acknowledgments

We thank John Platt for his assistance.